\documentclass[conference]{IEEEtran}
\IEEEoverridecommandlockouts
\usepackage{cite}
\usepackage{amsmath,amssymb,amsfonts}
\usepackage{algorithmic}
\usepackage{graphicx}
\usepackage{textcomp}
\usepackage{xcolor}
\def\BibTeX{{\rm B\kern-.05em{\sc i\kern-.025em b}\kern-.08em
    T\kern-.1667em\lower.7ex\hbox{E}\kern-.125emX}}
    
\usepackage{hyperref}
\usepackage{amsmath}
\usepackage{subcaption}

\definecolor{newcolor}{rgb}{.8,.349,.1}
    
\begin{document}

\title{Deep learning classification of bacteria clones explained by persistence homology
\thanks{This research was funded by the Priority Research Area Digiworld under the program Excellence Initiative – Research University at the Jagiellonian University in Kraków.}
}

\author{\IEEEauthorblockN{Adriana Borowa\IEEEauthorrefmark{1}\IEEEauthorrefmark{2}\IEEEauthorrefmark{5}, Dawid Rymarczyk\IEEEauthorrefmark{1}\IEEEauthorrefmark{2}\IEEEauthorrefmark{5}, Dorota Ocho\'nska\IEEEauthorrefmark{3}\IEEEauthorrefmark{4}, Monika Brzychczy-W\l{}och\IEEEauthorrefmark{3}\IEEEauthorrefmark{4} and Bartosz Zieli\'nski\IEEEauthorrefmark{1}\IEEEauthorrefmark{2}\IEEEauthorrefmark{4}} \IEEEauthorblockA{\IEEEauthorrefmark{1}Faculty of Mathematics and Computer Science, Jagiellonian University, Krak\'ow, Poland}
\IEEEauthorblockA{\IEEEauthorrefmark{2} Ardigen SA, Krak\'ow, Poland}\IEEEauthorblockA{\IEEEauthorrefmark{3} Department of Molecular Medical Microbiology, Chair of Microbiology,\\ Faculty of Medicine, Jagiellonian University Medical College, Krak\'ow, Poland}
\IEEEauthorblockA{\IEEEauthorrefmark{5} \textit{\{ada.borowa,dawid.rymarczyk\}@student.uj.edu.pl}}
\IEEEauthorblockA{\IEEEauthorrefmark{4} \textit{\{dorota.ochonska,m.brzychczy-wloch,bartosz.zielinski\}@uj.edu.pl}}
}
% Monika/Dorota
% Department of Molecular Medical Microbiology,
% Chair of Microbiology,
% Faculty of Medicine,
% Jagiellonian University Medical College

% \author{\IEEEauthorblockN{Adriana Borowa}
% \IEEEauthorblockA{
% \textit{Faculty of Mathematics} \\ 
% \textit{and Computer Science} \\
% \textit{Jagiellonian University}\\
% Krak\'ow, Poland \\
% \IEEEauthorblockA{\textit{Ardigen SA}} 
% Krak\'ow, Poland \\
% ada.borowa@student.uj.edu.pl}
% \and
% \IEEEauthorblockN{Dawid Rymarczyk}
% \IEEEauthorblockA{
% \textit{Faculty of Mathematics} \\ 
% \textit{and Computer Science} \\
% \textit{Jagiellonian University}\\
% Krak\'ow, Poland \\
% \IEEEauthorblockA{\textit{Ardigen SA}} 
% Krak\'ow, Poland \\
% dawid.rymarczyk@student.uj.edu.pl}
% \and
% \IEEEauthorblockN{Dorota Ocho\'nska}
% \IEEEauthorblockA{
% \textit{Chair of Microbiology}\\
% \textit{Faculty of Medicine}\\
% \textit{Jagiellonian University Medical College}\\
% Krak\'ow, Poland \\
% dorota.ochonska@uj.edu.pl}
% \and
% \IEEEauthorblockN{\hspace{2.5cm}Monika Brzychczy-W\l{}och}
% \IEEEauthorblockA{
% \hspace{2.5cm}\textit{Chair of Microbiology}\\
% \hspace{2.5cm}\textit{Faculty of Medicine}\\
% \hspace{2.5cm}\textit{Jagiellonian University Medical College}\\
% \hspace{2.5cm}Krak\'ow, Poland \\
% \hspace{2.5cm}m.brzychczy-wloch@uj.edu.pl}
% \and
% \IEEEauthorblockN{Bartosz  Zieli\'nski}
% \IEEEauthorblockA{
% \textit{Faculty of Mathematics} \\ 
% \textit{and Computer Science} \\
% \textit{Jagiellonian University}\\
% Krak\'ow, Poland \\
% \IEEEauthorblockA{\textit{Ardigen SA}} 
% Krak\'ow, Poland \\
% bartosz.zielinski@uj.edu.pl}}
\IEEEoverridecommandlockouts
\IEEEpubid{\makebox[\columnwidth]{978-0-7381-3366-9/21/\$31.00 ©2021 IEEE \hfill} \hspace{\columnsep}\makebox[\columnwidth]{ }}
\maketitle
% \IEEEpubidadjcol
\begin{abstract}
In this work, we automatically distinguish between different clones of the same bacteria species (Klebsiella pneumoniae) based only on microscopic images. It is a challenging task, previously seemed unreachable due to the high clones' similarity. For this purpose, we apply a multi-step algorithm with attention-based deep multiple instance learning, which returns parts of the image crucial to the prediction. Except for obtaining high accuracy, we introduce extensive explainability based on persistence homology, increasing the understandability and trust in the model. Our work opens a plethora of research pathways towards cheaper and faster epidemiological management.
\end{abstract}

%%%%%%%%%
\section{Introduction}
Distinguishing between bacteria clones is a crucial part of the epidemiological management because it helps to discover source of the infection in a hospital and to prevent new cases. There exist several techniques to differentiate between bacteria clones, such as ribotyping~\cite{Bratu2005}, amplified fragment length polymorphism~\cite{Jonas2004}, multilocus sequence typing~\cite{Diancourt2005}, and the most widely used pulsed-field gel electrophoresis~\cite{Han2013, pmid10618128, Mamlouk2006}. However, all of them require costly equipment and reagents, and are highly unstable. Therefore, they must be repeated multiple times to obtain reliable results.

\begin{figure}[t]
\centering
\includegraphics[width=0.75\linewidth]{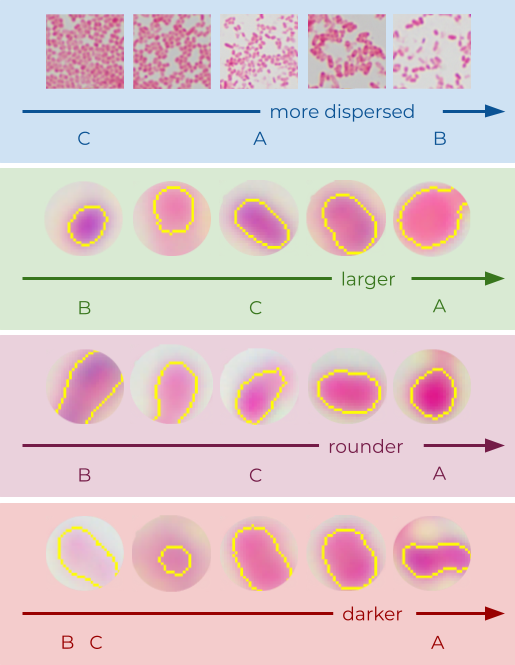}
\caption{Summary of statistically significant differences between the clones $A$, $B$, and $C$ of \textit{Klebsiella pneumoniae} based on the results presented in Section~\ref{sec:results}. Subsequent rows show differences in dispersion (blue arrows), size (green arrows), shape (purple arrows), and color (red arrows). Clone $A$ can be characterized as the largest, roundest, and darkest one - our method allows for description of distinct clones and finds features not visible to naked eye.}
\label{fig:properties}
\end{figure}

In this work, we automatically distinguish between different clones of the same bacteria species based only on microscopic images. This is an especially challenging task due to the high similarity of clones, as presented in Fig.~\ref{fig:clone_examples}, however, desirable due to acceleration of the diagnosis and reduction of the cost. To the best of our knowledge, it is the first attempt to solve this problem. Previous works either tried to use microscopic images to recognize bacteria morphology or classify different bacteria species~\cite{zielinski2017deep}, which is a much simpler task. As there exist no benchmarks we could use, we prepared an internal database containing $3$ clones with $31$ isolates of \textit{Klebsiella pneumoniae} (see Fig.~\ref{fig:clone_examples}), a Gram-negative rod-shaped bacteria that may cause pneumonia, urinary tract infections or even sepsis~\cite{Fung420, Karama2007, Ko2002}. 

Due to the large resolution of microscopic images, which cannot be scaled down due to loss of information, we first divide an image into patches, then generate their representations, and finally apply attention-based multiple instance learning, AbMILP~\cite{pmlr-v80-ilse18a}, to classify clones. AbMILP aggregates representations using the attention mechanism that promotes essential patches. Therefore, except for the classification results, it also returns the attention scores that correspond to the patches' importance. We also present results obtained using other techniques, like majority voting or mean and max pooling, but they either provide worse results or cannot be effectively explained.
At the same time, in AbMILP, patches with the highest scores can be used to explain results obtained for a specific image, but they do not provide systematic characteristics of the clones, neither from the spatial arrangement nor from the individual cell perspective. This characteristic, however, is necessary to trust the model because microbiologists commonly assume that differentiating the clones of the same bacteria type based only on microscopic images seems unreachable.

To effectively explain the AbMILP, we introduce two methods based on essential patches segmented using CellProfiler~\cite{carpenter2006cellprofiler}. The first one, based on persistence homology~\cite{edelsbrunner2000topological}, analyzes the spatial arrangement of the bacteria, while the second one examines the size, shape, and color of the individual cells, as presented in Fig.~\ref{fig:properties}. According to microbiologists, both XAI methods deliver a convincing explanation of the obtained results and increase the trust in the model.

The main contributions of this paper are as follows:
\begin{itemize}
\item Applying deep multiple instance learning to differentiate between bacteria clones based only on microscopic images (task previously seemed unreachable) with high accuracy.
% \item Introducing new post-hoc explainability method based on persistence homology that measures differences between the clones, such as the spatial arrangement of bacteria as well as their size, shape, and color.
% \item \todosb{Ada}{Usunac/zmienic?} Delivering a publicly available database containing $620$ images with $31$ isolates of \textit{Klebsiella pneumoniae}.
\item Introducing a new post-hoc explainability method based on persistence homology that measures differences in the spatial arrangement of cells between bacteria clones.
\item Providing systematic characteristic of the clones (their size, shape, and color) from the individual cell perspective.
\end{itemize}

%%%%%%%%%
\section{Related works}
\paragraph{Automatic pathogen classification based on microscopic images} Pathogen classification is a research problem addressed within different pathogens subgroups like bacteria~\cite{veropoulos1999automated, chayadevi2013extraction,  zielinski2017deep}, fungi~\cite{zhang2017automatic, tahir2018fungus, zielinski2019deep}, and  protozoa~\cite{widmer2002identification, castanon2007biological, kosov2018environmental}. In the case of bacteria, this topic was first examined in 1998 in ~\cite{veropoulos1999automated} where multi-layer perceptron was used to classify tuberculosis bacteria. Then, many approaches used hand-crafted features like perimeters, shapes, colors, contrast, and other morphological characteristics to distinguish between different bacteria species~\cite{liu2001cmeias,xiaojuan2009improved, kumar2010rapid, chayadevi2013extraction}. There were, as well, deep learning based approaches~\cite{nie2015deep, turra2017cnn, zielinski2017deep, wahid2018classification, rahmayuna2018pathogenic, ahmed2019combining}. Most of the work was focused on detecting only a specific species of the bacteria, e.g. tuberculosis bacteria~\cite{veropoulos1999automated, osman2010genetic, zhai2010automatic, rulaningtyas2011automatic, govindan2015automated, ghosh2016hybrid, priya2016automated, lopez2017automatic, panicker2018automatic, mithra2019automated}. Moreover, there was a lot of research in classifying the bacteria colonies. In~\cite{men2008application} authors used SVM, in~\cite{chen2009automated} one-class version of SVM with RBF kernel was used, and in~\cite{turra2017cnn} 2-layer CNN was employed. 
In summary, there was no research focused specifically on clone classification using microscopic images. And, even though there was some work on pathogen classification, none of methods used explainable AI to describe the rationale behind the prediction.

\paragraph{Post-hoc explainability in cells} There are only a few works touching on explainability of models for a cell classification. Among them is the research on the classification of the lung cancer~\cite{campanella2019clinical} where CNN network was used to create a patch image representation and  images were classified using bidirectional LSTM. The analysis of the LSTM gates showed that they are activated only when the patch contains cancer cells. Similar model was used in~\cite{graziani2018regression} but the explanations are based on the concept activation vectors~\cite{kim2018interpretability} adapted to the regression problem. The next example of XAI applications in histology is colorectal cancer researched in~\cite{sabol2020explainable} where cumulative fuzzy class membership criterion was used to explain the model decisions. Furthermore,~\cite{rymarczyk2020kernel} used self-attention and attention pooling maps to show the decisive parts of histology and fungi images, and \cite{brodzicki2020pre} used class activation maps to show the key features on the images that decided the classification of the \textit{Clostridioides difficile} bacteria cytotoxicity.

None of the previous works tried to utilize persistence homology to research dispersion characteristic of bacteria images. Moreover, according to best of our knowledge, there was no attempt to use Multiple Instance Learning to  to characterize biomedical images on the level of singular cells. 

%%%%%%%%%
\section{Material}

\begin{figure}[t]
\centering
\includegraphics[width=\linewidth]{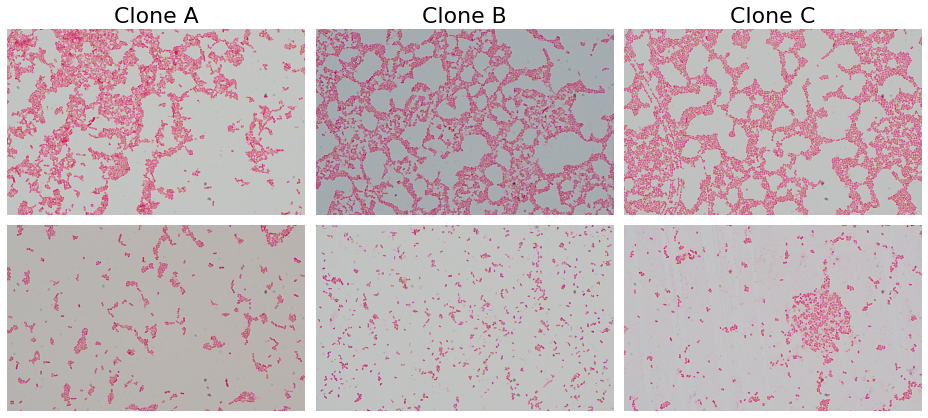}
\caption{Sample images from our database visualize that different clones of \textit{Klebsiella pneumoniae} are highly similar both from the spatial arrangement and from the individual cell perspective.}
\label{fig:clone_examples}
\end{figure}

\begin{figure*}[t]
\centering
\includegraphics[width=\linewidth]{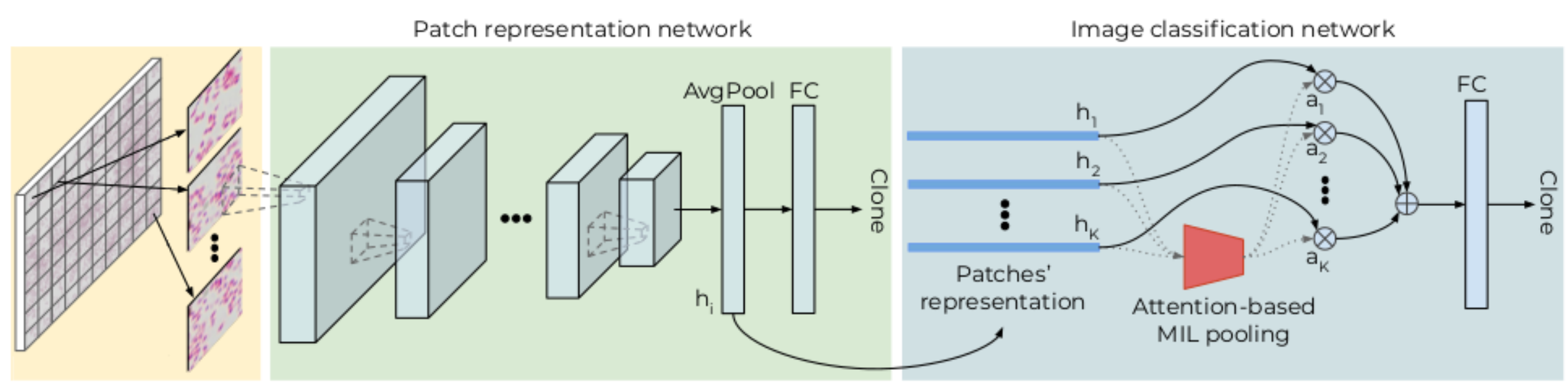}
\caption{The pipeline of the method contains four steps. First, the image is divided into patches that are preprocessed and passed through the representation network. Then, the patches' representations are aggregated with AbMILP to obtain one fixed-sized vector for each image. Finally, the vector is passed to the fully-connected layers of the network to classify the bacteria's clone. Representation and classification networks are trained separately due to the large number of patches per image. As a result of applying AbMILP, it is possible to obtain systematic characteristics of the clones.}
\label{fig:pipeline}
\end{figure*}

Our dataset contains $3$ clones of \textit{Klebsiella pneumoniae} further described as clones $A$, $B$, and $C$, containing $11$, $10$, and $10$ isolates, respectively. Each isolate is represented by a set of $20$ images from $2$ preparations ($10$ images per preparation) to minimize the environmental bias. Isolates were obtained from patients of one hospital and come from the respiratory tract, urine, wound swabs, blood, fecal samples, catheter JJ, and urethral swab. All of the samples were Gram-stained and analyzed using an Olympus XC31 Upright Biological Microscope equipped with an Olympus SC30 camera to present Gram-negative organisms. They were photographed using Olympus D72 with $12$ bit image depth and saved as TIFF images of size $3500\times5760$ pixels. Sample images from our database are presented in Fig.~\ref{fig:clone_examples}.

To cluster considered isolates into the clones, they were analyzed using pulsed-field gel electrophoresis (PFGE), a gold standard that relies on separating the DNA fragments after restriction cutting. PFGE profiles of the isolates were then interpreted using GelCompar software\footnote{\url{https://www.applied-maths.com/gelcompar-ii}} to obtain a dendrogram calculated with the Dice (band-based similarity) coefficient which is commonly used by microbiologists~\cite{REMENTERIA2001331}. According to the dendrogram (partially presented in supplementary materials\footnote{\url{http://ww2.ii.uj.edu.pl/~zielinsb/pdfs/clones_classification_sm.pdf}}) isolates cluster into three clones. Clones $A$ and $B$ are similar (Dice score of $92\%$), while clone $C$ is more different from the others (Dice score  of $77\%$). We use clones returned by GelCompar as the ground truth.

%%%%%%%%%
\section{Method}

We present the multi-step pipeline of our method in Fig.~\ref{fig:pipeline}. Firstly, the image is divided into patches (defined below). Then, the foreground patches are preprocessed and passed through the patches representation network. Obtained representations are then aggregated using attention-based multiple instance learning~\cite{pmlr-v80-ilse18a} to obtain one fixed-sized vector for each image. Finally, such a vector can be passed to the successive fully-connected (FC) layers of the network to classify the bacteria's clone. Due to the large number of patches per image, it is impossible to train the model at once. Therefore, we divided it into separately trained representation and classification networks using binary cross-entropy loss in both cases. They are described below, together with the preprocessing step.

\paragraph{Generating patches and preprocessing}
Each image is scaled by a factor of $0.5$ and divided into the patches of resolution $250\times250$ pixels using sliding window with stride $125$ (the optimal scale, resolution, and stride were obtained from preliminary experiments). To eliminate patches that are either empty or too populated with overlapping bacteria, we calculate standard deviation of the intensity of each patch, and reject patches with values lesser than $1.5$ (empty patches) and greater than $30$ (populous patches). The standard deviation thresholds were obtained experimentally. The remaining patches (most likely with non-overlapping bacteria) are normalized using the mean and the standard deviation computed based on $10000$ random training patches.

\paragraph{Patch representation network}
Normalized patches are passed through ResNet-18 network~\cite{DBLP:journals/corr/HeZRS15} pretrained on ImageNet and then finetuned with the last layer containing three neurons corresponding to three considered clones (according to the preliminary experiment, a version without finetuning resulted in poor accuracy). The representations of the patches are obtained by taking the output of the penultimate layer.

\paragraph{Image classification network}
To aggregate patches' representations of an image, we apply attention-based multiple instance learning, AbMILP~\cite{pmlr-v80-ilse18a,rymarczyk2020kernel}. It is a type of a weighted average pooling, in which the neural network determines weights of representations. More formally, if $\mathbf{h} = \{\mathbf{h_i} \in \mathbb{R}^L\}_{i=1}^N$ is the set of patches' representations of an image, then the output of the operator is defined as:
\begin{equation}\label{AbMIL_z}
\mathbf{z} = \sum_{i=1}^{N}a_i\mathbf{h_i},
\textup{ where }
a_i = \frac{\exp\left(\mathbf{w}^Ttanh(\mathbf{Vh_i})\right)}{\sum_j^N\exp\left(\mathbf{w}^Ttanh(\mathbf{Vh_j})\right)},
\end{equation}
with $\mathbf{w} \in \mathbb{R}^{M}$ and $\mathbf{V} \in \mathbb{R}^{M \times L}$ being trainable layers of the neural network. Aggregated vector $\mathbf{z}$ is further processed by a dense layer with three output neurons corresponding to three considered clones.

\begin{figure*}[ht]
\centering
\includegraphics[width=\linewidth]{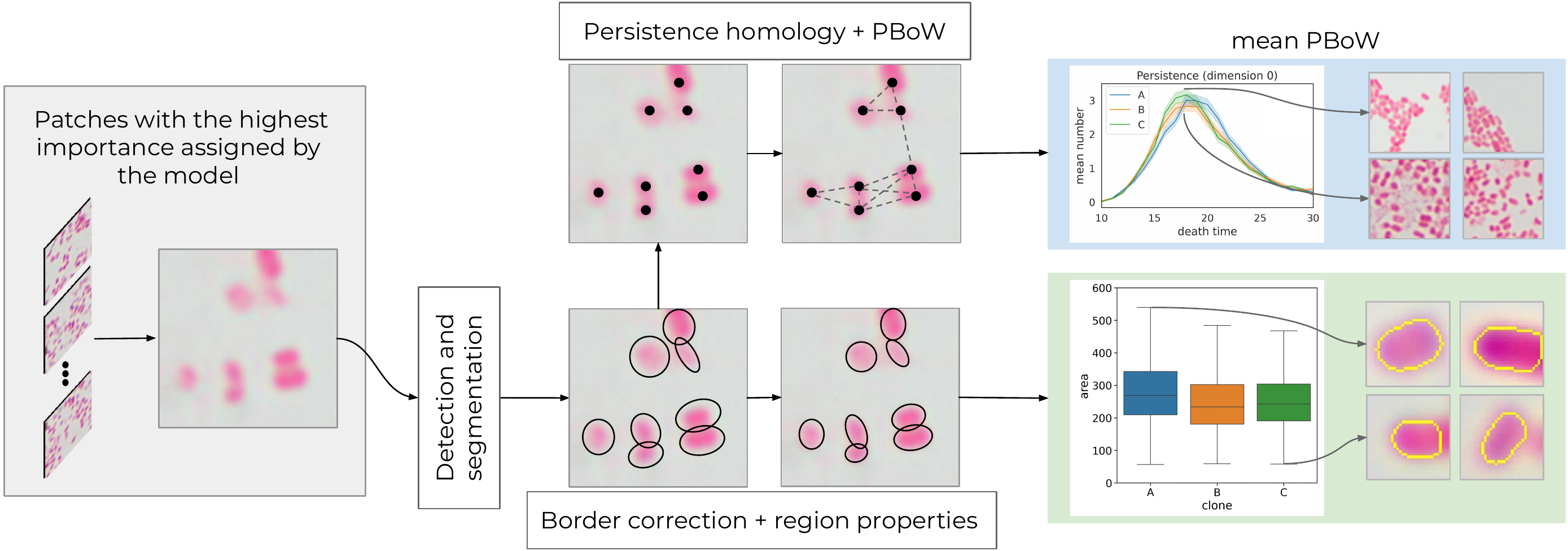}
\caption{After training the model, we obtain patches that have the highest importance assigned by that model. We perform detection of cells on the patch and their segmentation. Then, pipeline divides into two paths. In the first one, we find centers of cells, we generate point cloud that is analyzed using persistence homology, and finally we obtain information about dispersion of cells in each clone. The second path consists of the border correction and finding region properties of the cells, resulting in information about the mean size, shape, and color of cells in each clone.}
\label{fig:interpretability_pipeline}
\end{figure*}

%%%%%%%%%
\section{Explainability}
\label{sec:interpret}

To explain which properties of bacteria are important for the AbMILP when distinguishing between two clones, we create a set of important patches that contains two patches with the highest attention weight ($a_i$ value) from each correctly predicted image. We decided to analyze such a set because it includes patches with the greatest influence on the results.

In Fig.~\ref{fig:interpretability_pipeline}, we present two types of the analysis conducted on such essential patches. The first one, based on persistence homology~\cite{edelsbrunner2000topological}, analyzes the spatial arrangement of the bacteria, while the second one examines the size, shape, and color of the individual cells. Both types of the analysis are preceded by bacteria cells' segmentation obtained using CellProfiler~\cite{carpenter2006cellprofiler} with a pipeline describe below.

\paragraph{Segmentation of bacteria}
To segment bacteria, we first transform a patch into the grayscale images and then apply a CellProfiler pipeline with only one component identifying primary objects with a typical diameter between $10$ and $40$ pixels\footnote{For details, see \url{http://cellprofiler-manual.s3.amazonaws.com/CellProfiler-3.0.0/modules/objectprocessing.html\#identifyprimaryobjects}}. The objects with diameters outside this range are discarded along with objects touching the border. The result of segmentation for sample patches is presented in supplementary materials.
    
\paragraph{Spatial arrangement of bacteria}
For each segmentation of a patch, we generate a point cloud $C \subset \mathbb{R}^2$ containing centers of the segmented bacteria and analyze this point cloud using persistence homology~\cite{edelsbrunner2000topological}, that provides a comprehensive, multiscale summary of the underlying data's shape and is currently gaining increasing importance in data science~\cite{ferri2017persistent}.

Persistence homology can be defined for a continuous function $f :C \rightarrow \mathbb{R}$, typically a distance function from a collection of points (in this case, centers of the bacteria). Focusing on sub-level sets $L_c = f^{-1}( (-\infty,c] )$, we let $c$ grow from $-\infty$ to $+\infty$. While this happens, we can observe a hierarchy of events. In dimension $0$, connected components of $L_c$ are created and merged. In dimension $1$, cycles that are not bounded or higher dimensional voids appear in $L_c$ at critical points of $f$. The value of $c$ on which a connected component, cycle or a higher dimensional void, appears is referred to as \emph{birth time}. They subsequently either become identical (up to a deformation) to other cycles and voids (created earlier) or glued-in and become trivial. The value of $c$ on which that happens is referred to as \emph{death time}. As a result, every connected component, a cycle, or a higher dimensional void can be characterized by a pair of numbers, $b$ and $d$, the birth and death time. Therefore, a set of birth-death pairs is obtained, called a persistence diagram or a persistence barcode. They are not easy to compare due to their variable size. Therefore, we apply a vectorization method called persistence bag of words~\cite{zielinski2019persistence,zielinski2020persistence} to obtain a fixed-size representation for each patch. Based on them, we calculate the average persistence bag of words for each clone and analyze differences in representative patches. for more information about persistent homology please see \cite{edelsbrunner2010computational}.

\paragraph{Properties of individual bacteria}
To understand what differentiates individual cells between the clones, we limit segments obtained from CellProfiler to those isolated from the others (with no adjacent segments). As the boundaries of cells returned by CellProfiler are not ideal, we correct them using Otsu thresholding on a dilated segment. Then, we fit an ellipse to the corrected segment and analyze significant differences between clones, such as distributions of size, shape (minor to major axis length), and color. See supplementary materials for details.

%%%%%%%%%
\section{Experiment setup}
\label{sec:experiment_setup}

We train the representation network and classification network $5$ time using cross-validation to obtain reliable results for each configuration. To create a training and test set for each fold, we randomly choose one of two isolate's preparations and add it to the training set. The remaining preparation is added to the test set.

\begin{table*}[t]
\renewcommand{\arraystretch}{1.3}
    \captionsetup{justification=centering, labelsep=newline}
    \caption{\footnotesize{\textsc{Results of Image-based Classification for Seven Tested Methods With the Best Scores in Bold Based on Wilcoxon Signed-rank Test.}}}
    % \caption{Results of image-based classification for seven tested methods with the best scores in bold. One can observe that AbMILP achieves the best results in all metrics. Notice that Wilcoxon signed-rank test is performed between the best method (the best mean accuracy) and all the others to identify which significantly differ from the best-obtained result and which ones do not and can be considered equally good.}
    \label{tab:all_results}
    \centering
    \begin{tabular}{|c|c|c|c|c|c|}
    \hline
        \textbf{Method} & \textbf{Accuracy} & \textbf{Precision} & \textbf{Recall} & \textbf{F1} & \textbf{AUC} \\
        \hline
        % \hline
        instance + mv & $\mathbf{63.8\pm1.1}$ & $64.1\pm1.0$ & $\mathbf{63.8\pm1.1}$ & $\mathbf{63.8\pm1.0}$ &$n/a$ \\
        \hline
        instance + max & $59.3\pm3.6$ & $59.8\pm3.5$ & $59.3\pm3.6$ & $59.3\pm3.6$ & $75.0\pm2.5$\\
        \hline
        instance + mean & $59.7\pm2.7$ & $60.5\pm2.7$ & $59.7\pm2.7$ & $59.8\pm2.7$ & $\mathbf{78.6\pm2.5}$\\ 
        \hline
        embedding + max & $56.3\pm6.9$ & $52.9\pm11.0$ & $56.3\pm6.9$ & $52.9\pm9.9$ & $68.3\pm9.6$\\
        \hline
        embedding + mean & $58.4\pm5.9$ & $58.9\pm6.3$ & $58.4\pm5.9$ & $\mathbf{58.0\pm6.2}$ & $73.4\pm5.1$\\
        \hline
        embedding + AbMILP & $\mathbf{64.8\pm2.2}$ & $\mathbf{66.9\pm1.4}$ & $\mathbf{64.8\pm2.2}$ & $\mathbf{64.3\pm2.4}$ & $\mathbf{77.3\pm2.7}$\\
        \hline
        embedding + SA-AbMILP & $62.7\pm2.0$ & $63.5\pm1.8$ & $62.7\pm2.0$ & $62.3\pm2.2$ & $\mathbf{78.1\pm1.9}$\\
        \hline
    \end{tabular}
\end{table*}

The representation network is finetuned with batch size $64$ and initial learning rate $0.0001$, decreasing $10$ times after every $1000$ steps. Hyperparameters were obtained during preliminary experiments which included different constant learning rates in range [$0.00000001, 0.001$] and batch size values: $16, 32, 64$. In these experiment we used only the training set. All network layers are finetuned with binary cross-entropy loss and weighted sampling to reduce the influence of data imbalance and were trained for $5000$ iterations, up to a point of flattening of the loss function. Training images were augmented using color jittering, random rotation, and random flip.

To obtain the best hyperparameters for the classification network we applied grid search over learning rate in the range [$0.000005$, $0.001$] and weight decay in the range [$0.00001$, $0.05$]. We use a standard number of attention heads ($3$) as according to~\cite{pmlr-v80-ilse18a} this parameter is not relevant.

When it comes to explainability, we build the Vietoris-Rips complex from the distance matrix, based on which we then calculate persistence homology of dimension $0$ using the Gudhi library~\cite{maria2014gudhi}. As a result, we obtain persistence diagrams with birth time equal to $0$ in all its points. Based on them, we generate a persistence bag of words with $128$ regular codewords of size $1$ with cluster centers in $0.5, 1.5, 2.5, \dots, 127.5$ of death time.

\begin{figure}[t]
\centering
\begin{subfigure}[b]{0.49\linewidth}
  \centering
  \includegraphics[width=\linewidth]{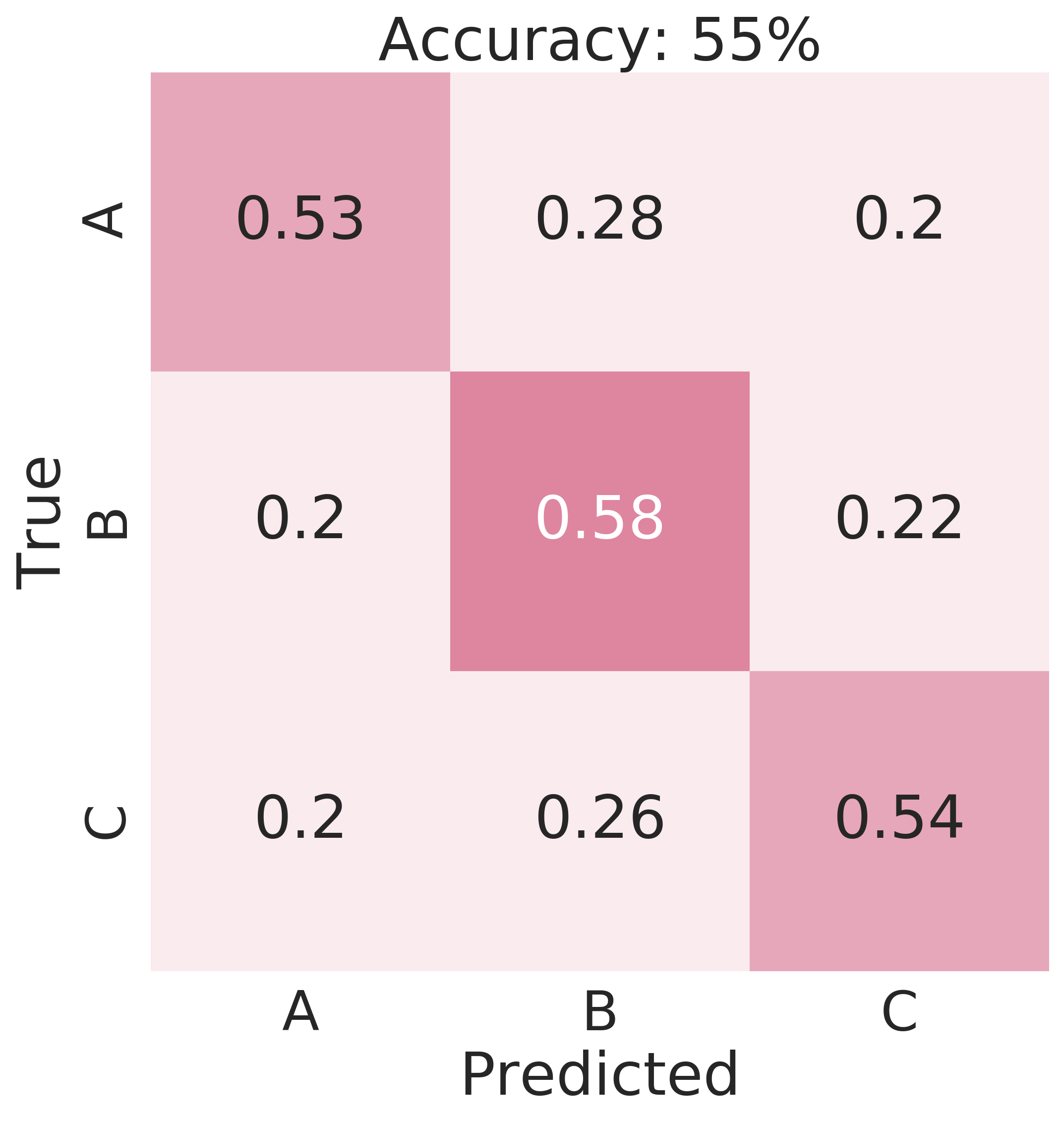}
  \caption{CNN}
\end{subfigure}
\begin{subfigure}[b]{0.49\linewidth}
  \centering
  \includegraphics[width=\linewidth]{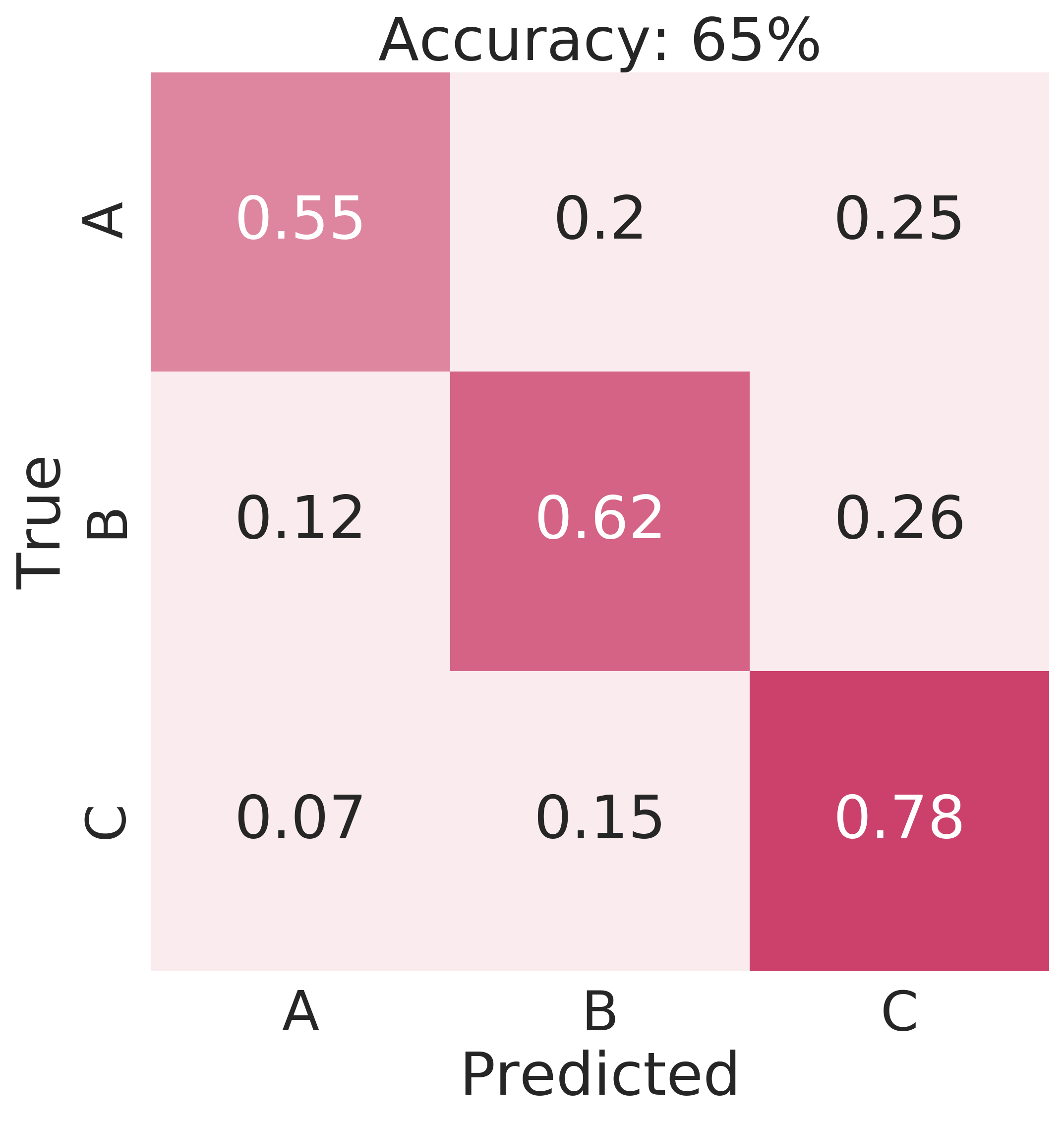}
  \caption{AbMILP}
\end{subfigure}
\caption{Confusion matrices for representation (CNN) and classification (AbMILP) networks averaged over $5$ folds and calculated on test subset. Usage of attention-based pooling visibly improves accuracy.}
\label{fig:cm}
\end{figure}

\begin{figure}[t]
\centering
\begin{subfigure}[b]{0.85\linewidth}
  \centering
  \includegraphics[width=\linewidth]{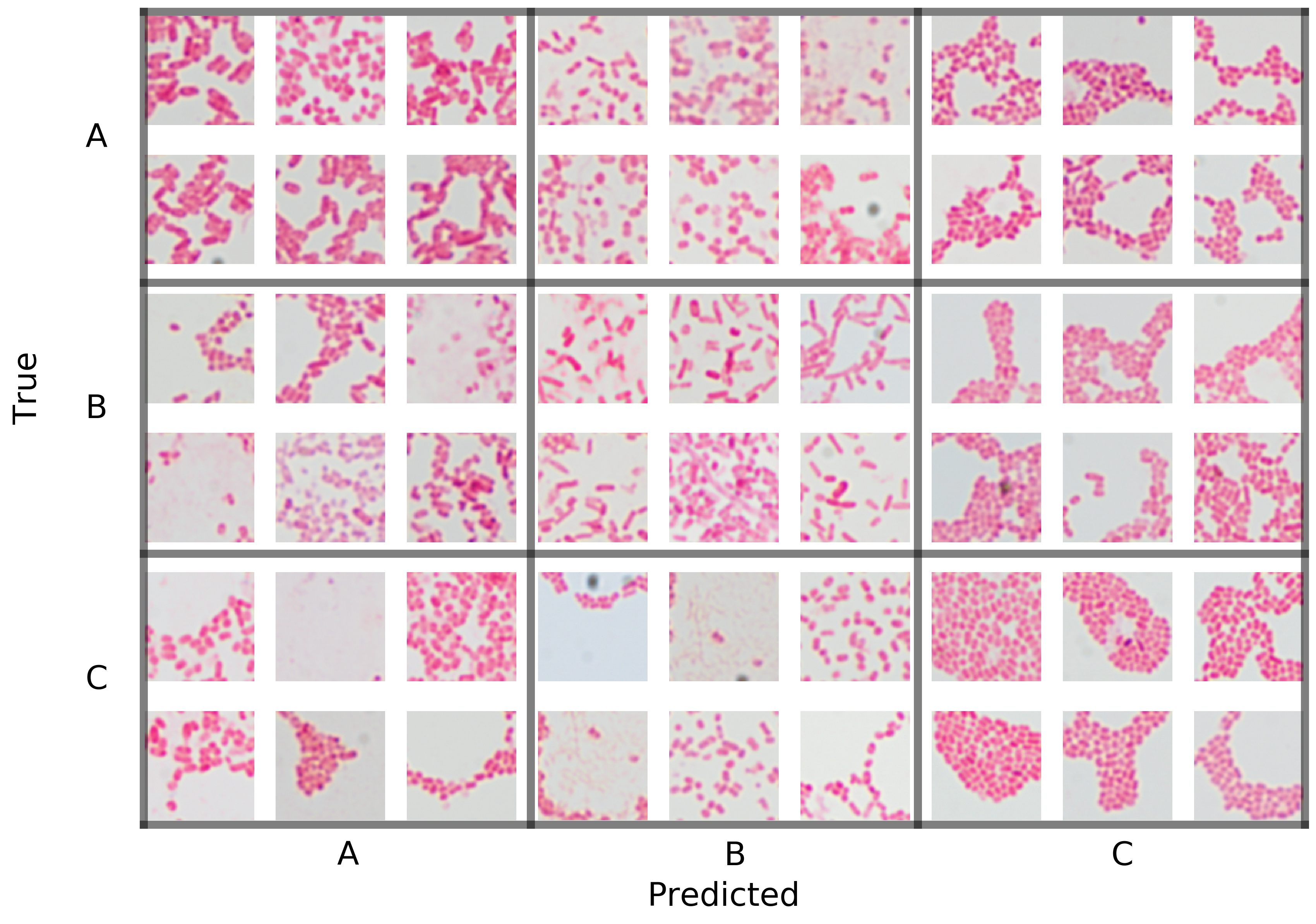}
  \caption{CNN}
\end{subfigure}
\begin{subfigure}[b]{0.85\linewidth}
  \centering
  \includegraphics[width=\linewidth]{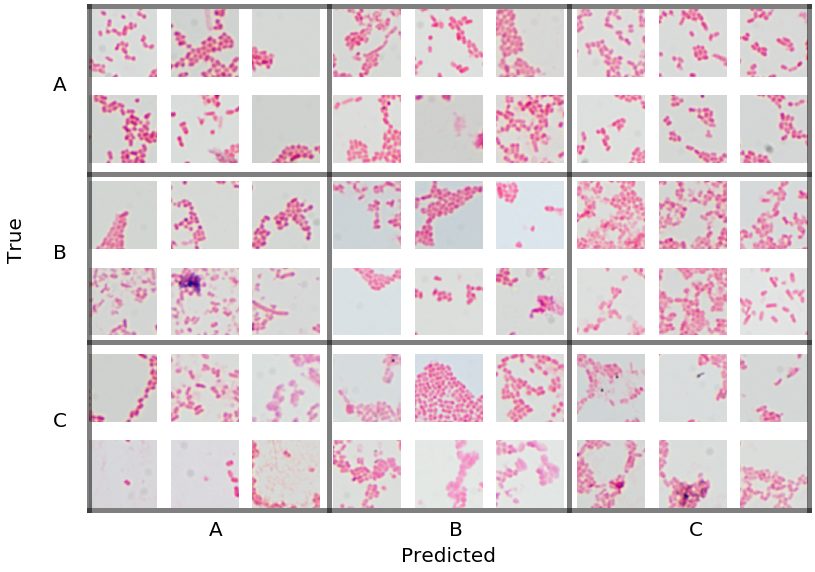}
  \caption{AbMILP}
\end{subfigure}
\caption{Qualitative confusion matrices for representation (CNN) and classification (AbMILP) networks, for fold $0$ of test subset. Each confusion matrix cell contains $6$ ($2 \times 3$) representative patches coming from different images. We cannot notice any unique characteristic of clones based on the patches from AbMILP. Thus, we introduce methods described in Sec.~\ref{sec:interpret} to find those characteristics.}
\label{fig:top_patches}
\end{figure}

We test variety of pooling methods to perform per image classification, because per patch results given by using only the CNN were not satisfactory with $54.9\%\pm2.2\%$ accuracy. Methods used to obtain per image results:
\begin{itemize}
    \item instance + mv: majority voting performed on all the patches in the image.
    \item instance + max: max pooling performed on representation network output.
    \item instance + mean: mean pooling performed on representation network output.
    \item embedding + max: max pooling performed on embeddings obtained from penultimate layer of representation network.
    \item embedding + mean: mean pooling performed on embeddings obtained from penultimate layer of representation network.
    \item embedding + AbMILP: attention based pooling performed on embeddings obtained from penultimate layer of representation network.
    \item embedding + SA-AbMILP: attention based pooling with self-attention kernel~\cite{rymarczyk2020kernel} performed on embeddings obtained from penultimate layer of representation network.
\end{itemize}

We perform all the experiments on a workstation with two $23$ GB GPU and $128$ GB RAM. On average, it takes $30$ hours to train a patch representation network and $8$ hours to generate patch representations for the classification step. Training of image classification network lasts up to $1$ hour. Both networks were implemented using PyTorch. The code is publicly available.\footnote{\url{https://github.com/gmum/clones_classification}}

%%%%%%%%%
\begin{figure}[t]
\centering
\includegraphics[width=\linewidth]{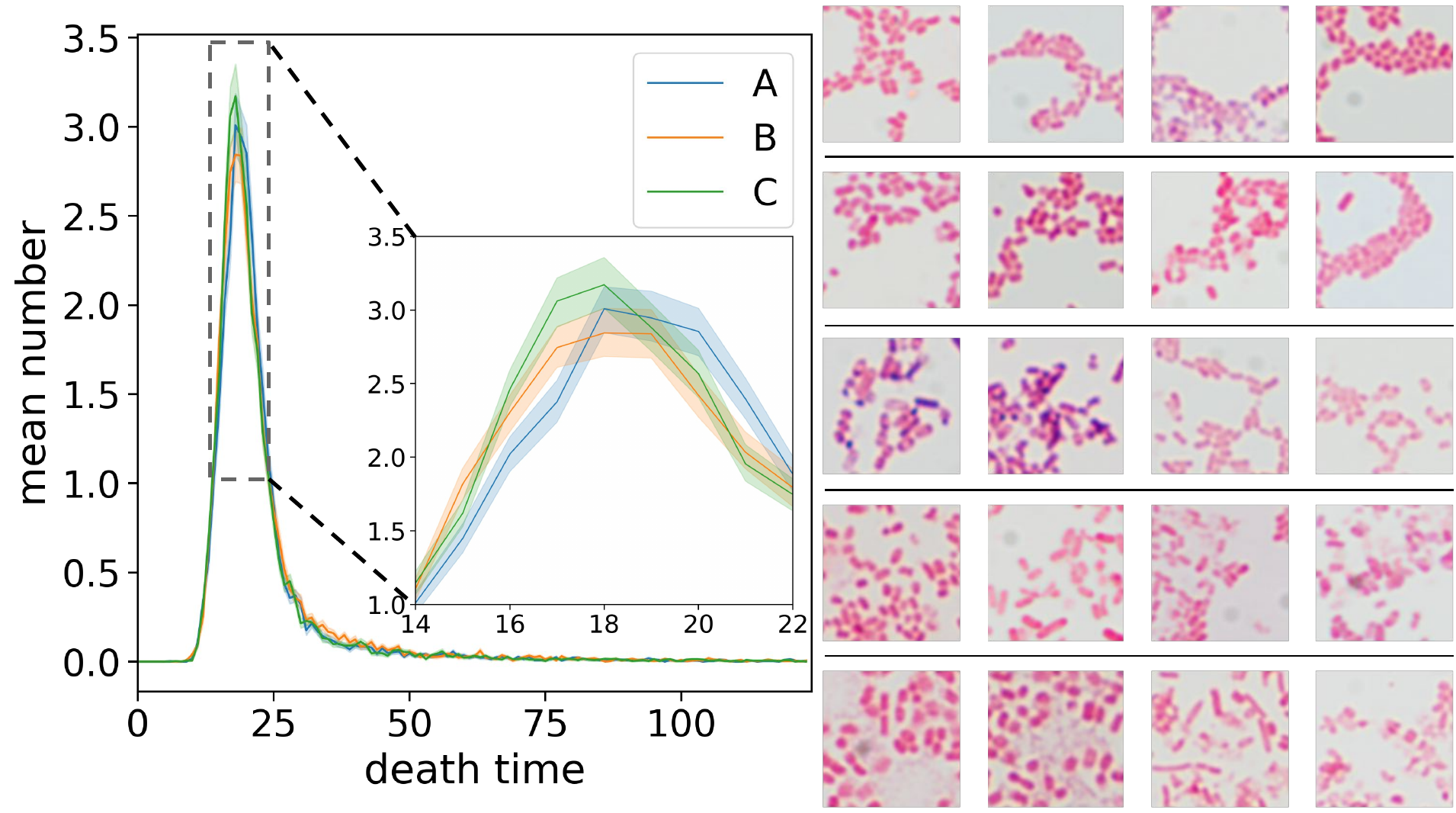}
\caption{The average persistence bag of words for clone C is significantly higher than clone B in the $[14; 22]$ death time interval. It means that many clone C bacteria are localized close to each other (the upper row of patches) and merge faster than clone B bacteria that are more dispersed (the bottom row of patches). It is one of the many conclusions drawn by the explainability framework based on the persistence homology.}
\label{fig:persist}
\end{figure}
\section{Results}
\label{sec:results}
\subsection{Classification}
\label{sec:results_classification}
Results are presented in Fig.~\ref{fig:cm} in the form of confusion matrices averaged over five folds for patch representation and image classification networks. They clearly indicate that it is possible to differentiate between bacteria clones based only on microscopic images with high accuracy. Moreover, while patch-based test accuracy can be relatively low at $55\%$, it can be compensated by attention-based pooling which obtains the accuracy of image-based classification at the level of $65\%$, whilst the accuracy of a random classifier is $33\%$. Confusion matrix of the image-based model indicates higher accuracy when classifying clone C. However, it is expected because it is more genetically distant to other clones and there are more similarities between clones $A$ and $B$. 

In Tab.~\ref{tab:all_results}, we present per-image results for the seven tested methods. The AbMILP method gives significantly better results, based on Wilcoxon signed-rank test, than other methods across all metrics and majority voting obtains almost as good results. However, in contrast to AbMILP and SA-AbMILP, majority voting is not useful method when trying to explain the model. Most probably, AbMILP achieves the best results because it is an approach fully parameterized by neural networks, which uses information from each of the instances in a weighted manner. It is surprising that AbMILP obtains better results than SA-AbMILP, nonetheless, we did not test all possible kernels in the later.

%%%%%%%%%
\subsection{Explainability}
\label{sec:results_interpret}

As the first type of a model explanation, we present quantitative confusion matrices obtained for test set by representation and classification networks (see Fig.~\ref{fig:top_patches}). While some trends are visible in representation networks, such as the more dispersed distribution of clone $B$, no systematic characteristic can be implicated in AbMILP, which demonstrates the need to explain it with more sophisticated explainability methods.

\begin{figure}[t]
\centering
\begin{subfigure}[b]{\linewidth}
  \centering
  \includegraphics[width=\linewidth]{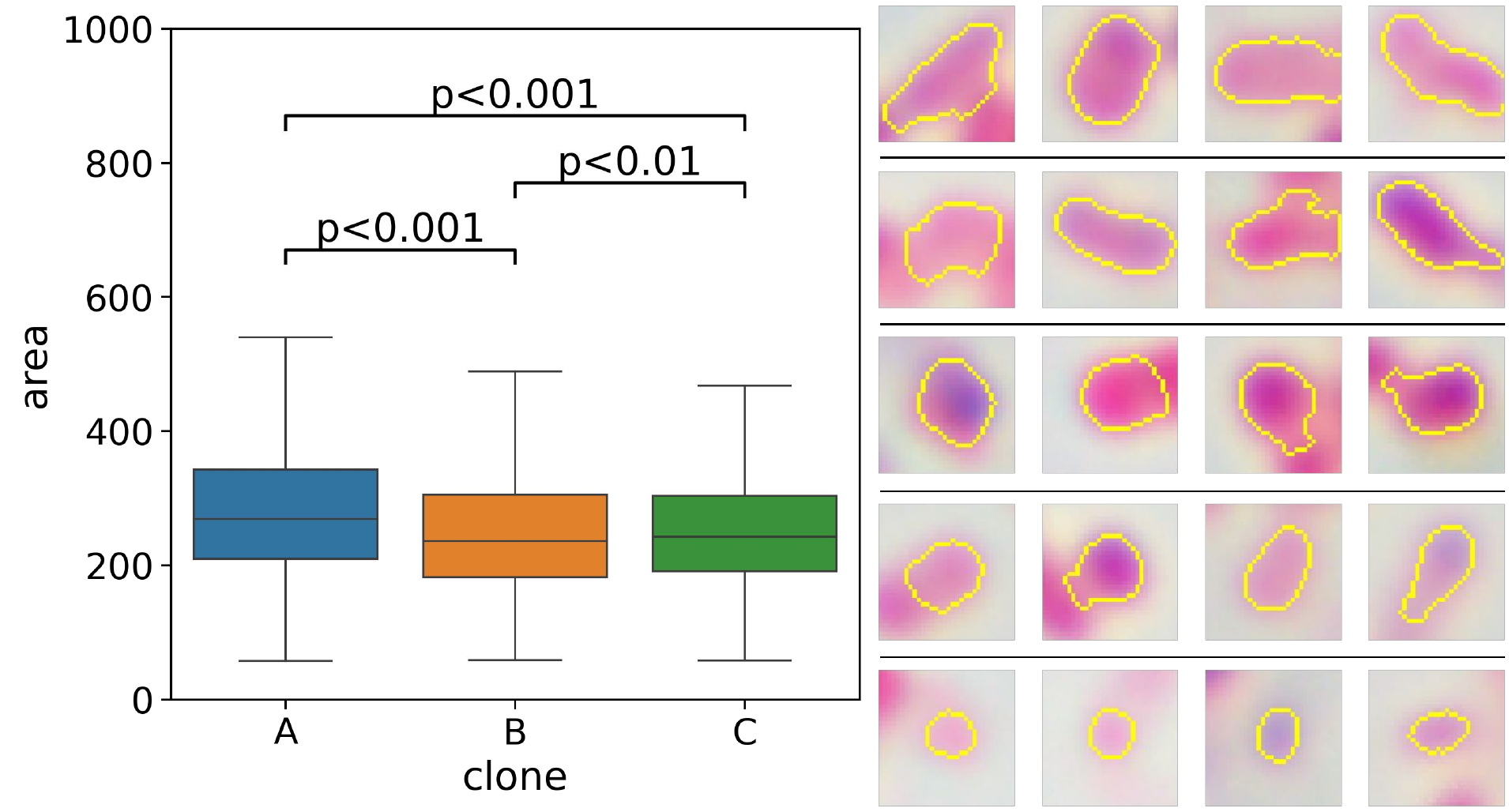}
  \caption{distribution of size}
\end{subfigure}
\begin{subfigure}[b]{\linewidth}
  \centering
  \includegraphics[width=\linewidth]{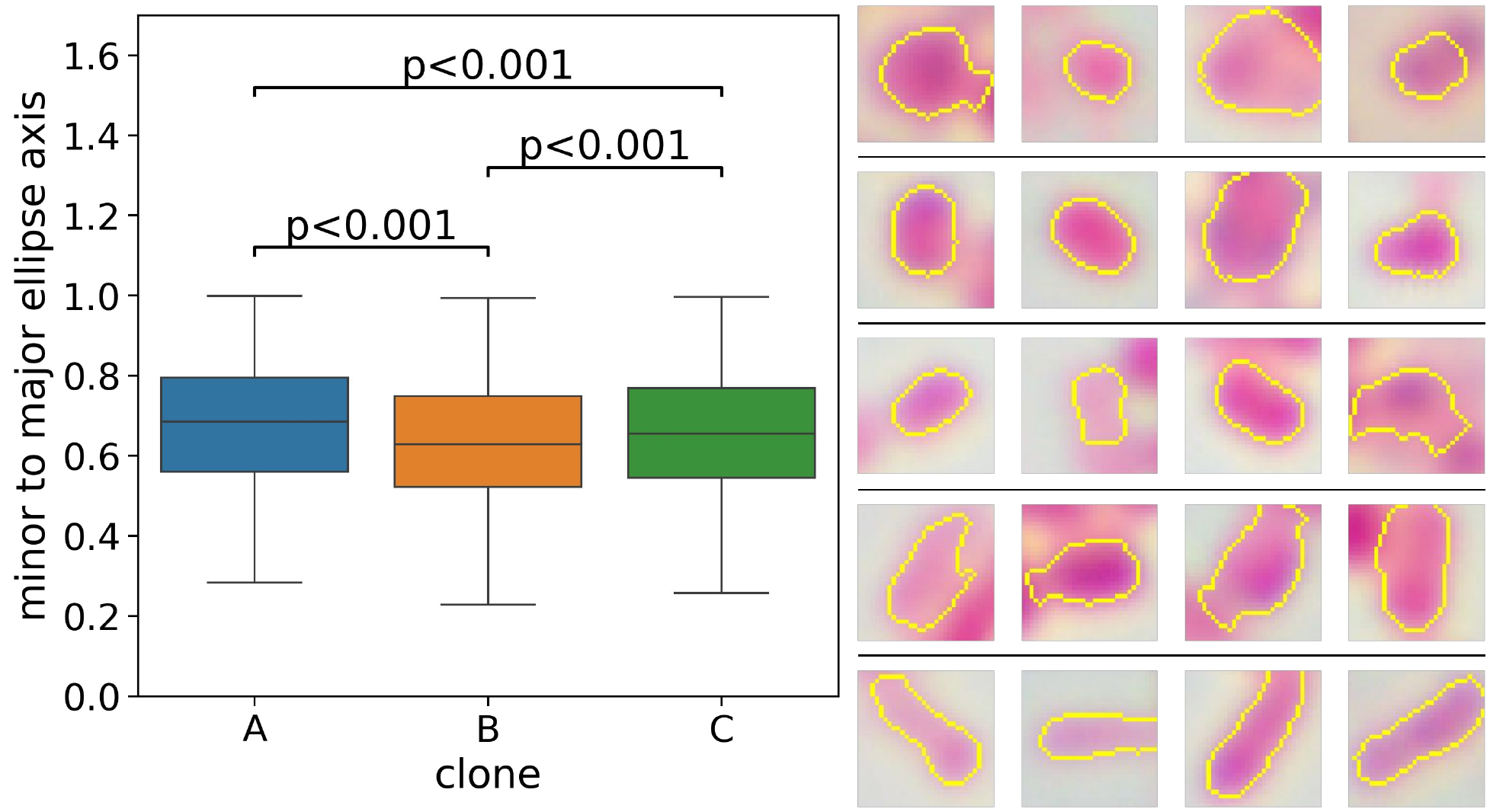}
  \caption{distribution of shape (roundness)}
\end{subfigure}
\begin{subfigure}[b]{\linewidth}
  \centering
  \includegraphics[width=\linewidth]{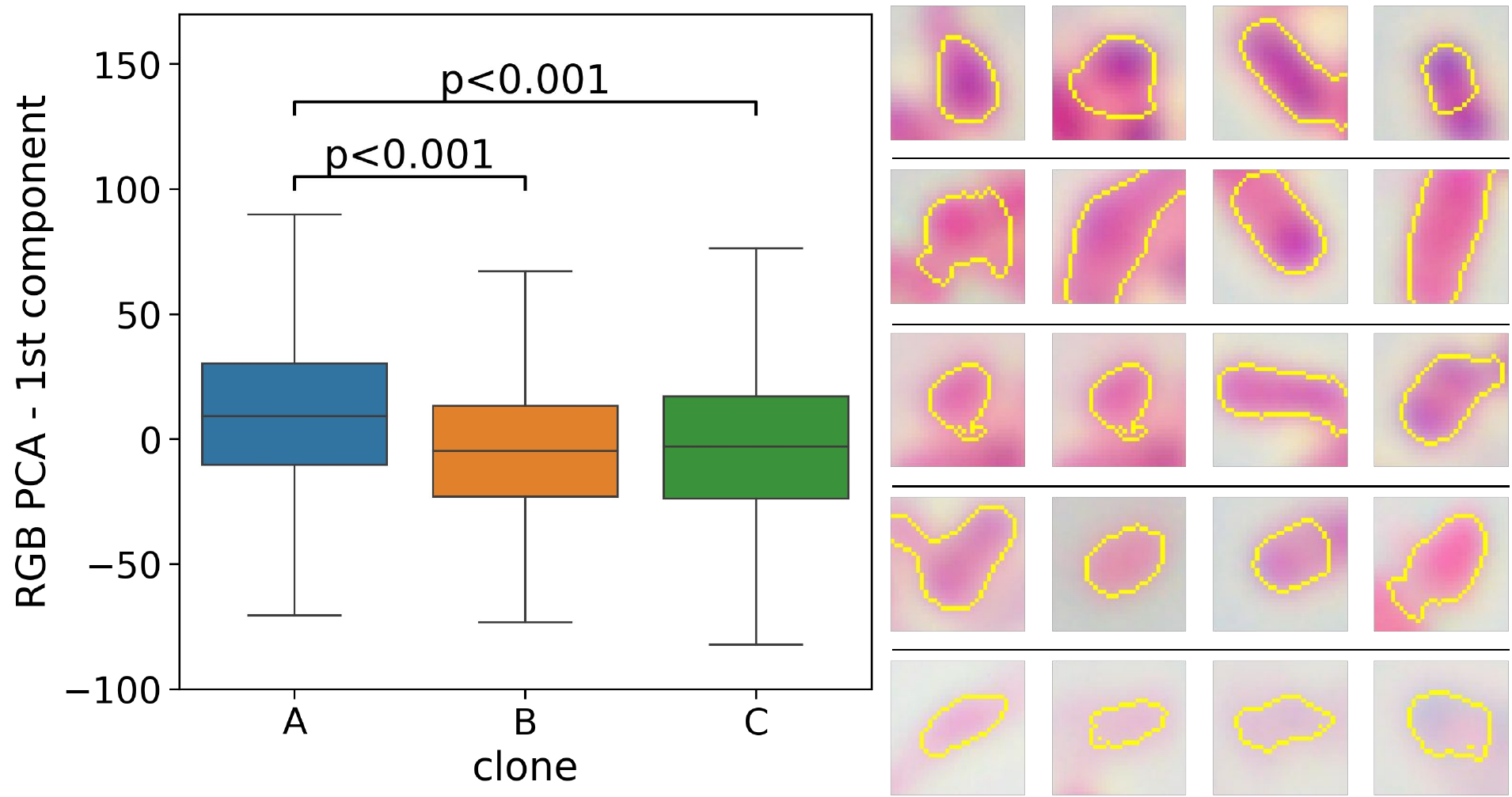}
  \caption{distribution of color intensity}
\end{subfigure}
\caption{Distributions of size (a), shape (b), and color intensity (c), as well as sample bacteria cells from the lowest (left side) to the highest value (right side) of considered properties. In all cases differences between distributions are statistically significant and the only exception is the distribution of color intensity between clones $B$ and $C$. }
\label{fig:individual_statistics}
\end{figure}

The first method is based on the persistence bag of words (PBoW) generated for the point clouds obtained for all important patches. In Fig.~\ref{fig:persist}, we present the average PBoW for all three clones that can be used to draw many interesting conclusions. Firstly, the average PBoW differs mostly between 14 and 22 death time and is the highest for clone C. This means that clone C contains much more cells merging at the initial stage of filtration. Hence, its cells are much closer to each other than in the case of clones A and B. Moreover, while clone B and C are in line in the interval [14; 22], although B is generally lower than C, the average PBoW for clone A is moved toward the right side of the plot. This means that clone B has either very close or very distant cells, while clone A contains rather slightly more distant cells than in the case of clone C.

In Fig.~\ref{fig:individual_statistics}  we present more individual properties of the bacteria, such as the larger size of bacterial cells from clones $A$ and $C$ comparing to clone $B$. One can also observe that bacterial cells of clone $A$ are larger and darker than in clones $B$ and $C$. Moreover, bacterial cells of clone $A$ are also significantly rounder than in clone $B$. Simultaneously, bacterial cells of clone $A$ are darker than in clones $B$ and $C$. We hypothesize that it can be caused by the presence of a polysaccharide capsule (envelope). More precisely, the thicker capsule results in darker color caused by dye retention during staining, rounder shape caused by the limited ability to reshape, and larger size due to dye accumulation both in the cell and in the envelope. We present a summary of the clones' properties in Fig.~\ref{fig:properties}.

%%%%%%%%%
\section{Conclusion}

In this work, we analyzed if it is possible to distinguish between different clones of the same bacteria species (\textit{Klebsiella pneumoniae}) based only on microscopic images. To address this challenging task, which previously seemed unreachable, we applied a multi-step algorithm with attention-based multiple instance learning (AbMILP), which resulted in an accuracy of $65\%$. Moreover, we pointed out the AbMILP interpretation's weaknesses, and we introduced more efficient explainability methods based on CellProfiler segmentation and persistence homology.

The extensive analysis revealed systematic differences between the clones from the spatial arrangement and the individual cell perspective. Hence, in the future, we plan to use this knowledge and cluster the isolates into clones using unsupervised methods to obtain cheaper and faster epidemiological management.

\bibliographystyle{IEEEtran}
% \bibliographystyle{model2-names.bst}
% \biboptions{authoryear}
\bibliography{bacteria_clones_wacv}

\end{document}

\title{Deep learning classification of bacteria clones explained by persistence homology -- supplementary materials
}

\author{\IEEEauthorblockN{Adriana Borowa} 
\and
\IEEEauthorblockN{Dawid Rymarczyk}
\and
\IEEEauthorblockN{Dorota Ocho\'nska}
\and
\IEEEauthorblockN{Monika Brzychczy-W\l{}och}
\and
\IEEEauthorblockN{Bartosz  Zieli\'nski}
}

\maketitle
\begin{figure}[ht]
\centering
\includegraphics[width=0.9\linewidth]{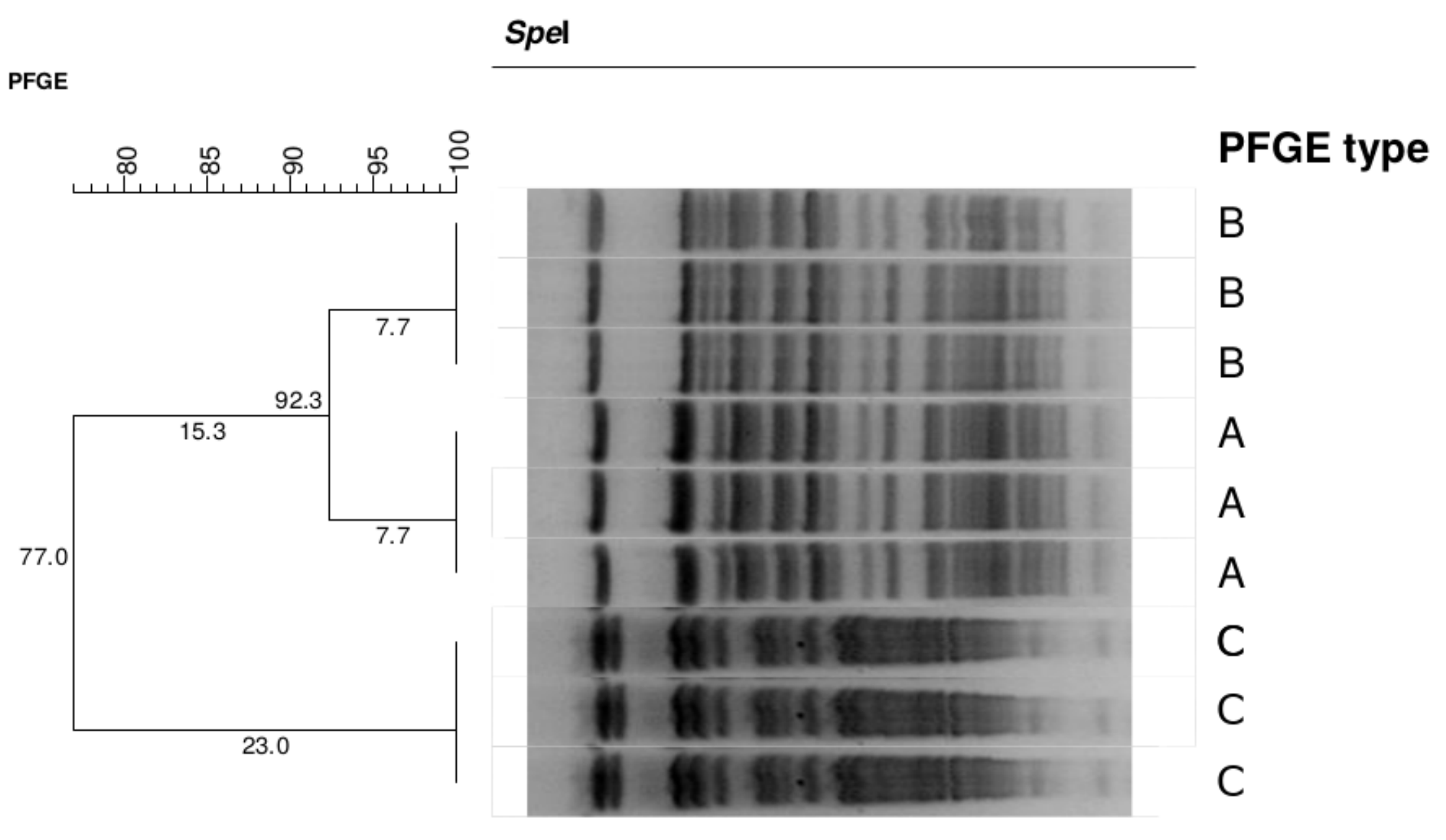}
\caption{Dendrogram calculated for $9$ isolates of \textit{Klebsiella pneumoniae} ($3$ per clone) based on GelCompar software. This figure limits the original number of $31$ isolates to $9$ for paper clarity. Dendrogram specifies the level of similarity between the clones, e.g. clones $A$ and $B$ are closer genetically to each other than to clone $C$.}
\label{fig:dendrogram}
\end{figure}
\begin{figure}[ht]
    \centering
    \includegraphics[width=\linewidth]{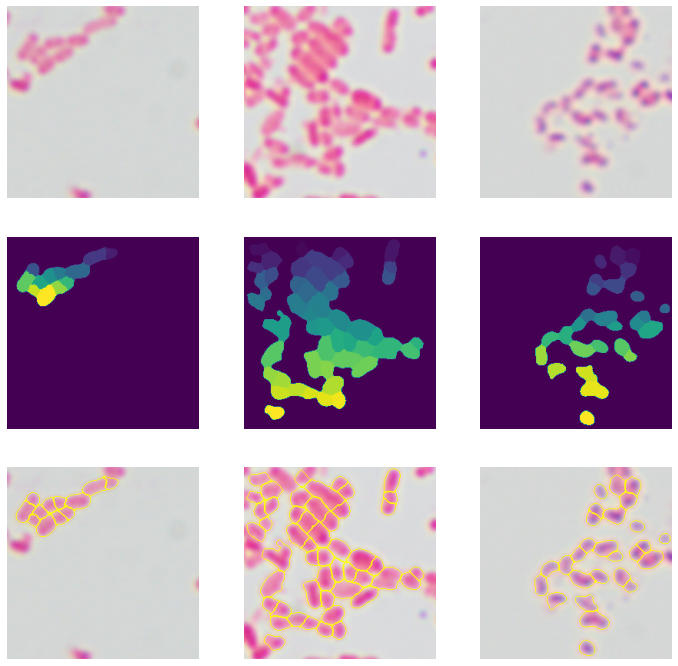}
    \caption{Segmentation for sample patches, as returned by the CellProfiler pipeline.}
    \label{fig:segmentation}
\end{figure}
\begin{figure}[ht]
    \centering
    \includegraphics[width=\linewidth]{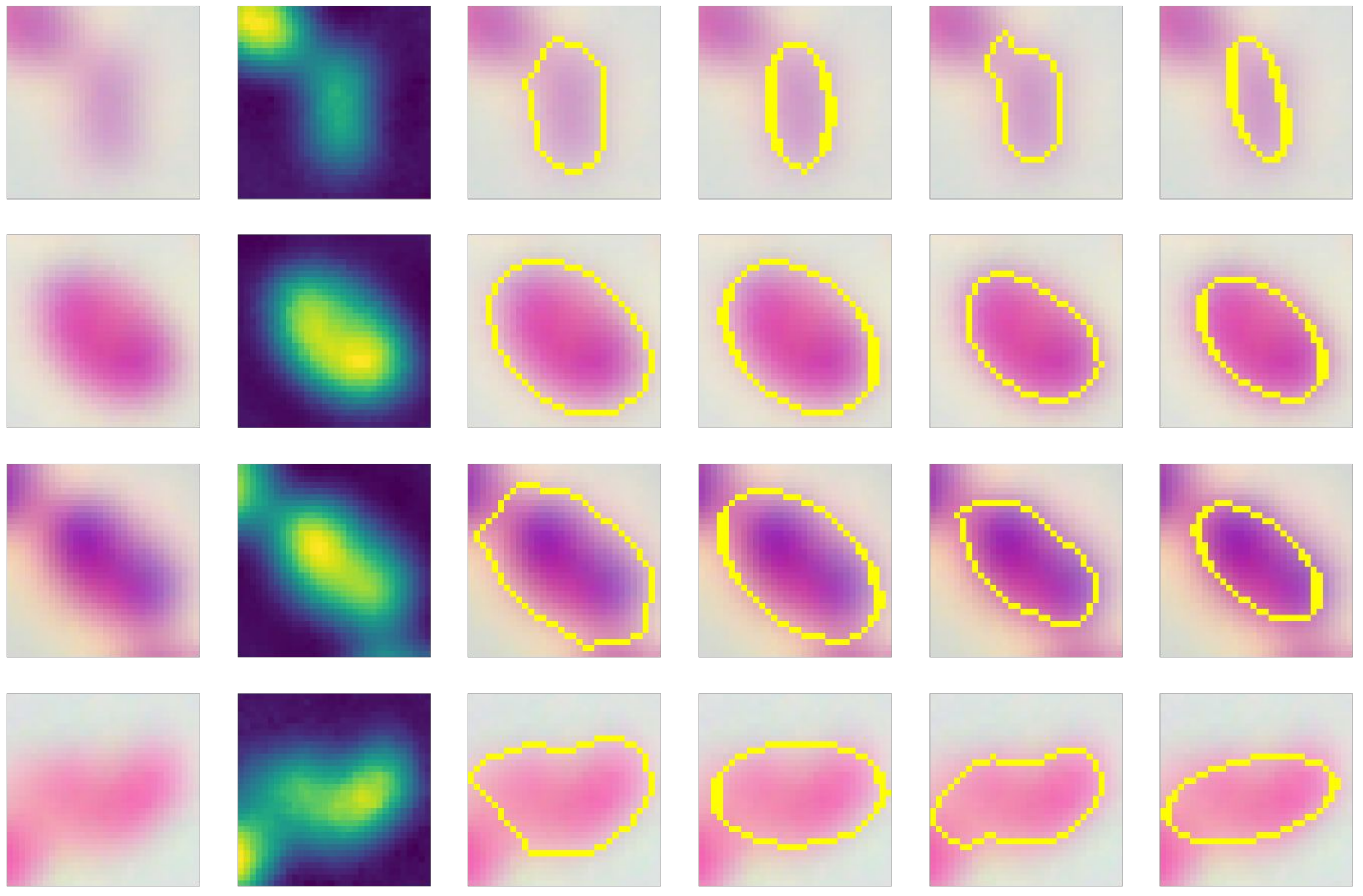}
    \caption{Border correction: firstly, we apply CellProfiler pipeline that finds isolated segments in the image. Then, to correct a segmentation, we apply Otsu thresholding on dilated segment and fit an ellipse.}
    \label{fig:border}
\end{figure}

In these supplementary materials, we present:
\begin{itemize}
    \item Figure~\ref{fig:dendrogram}: Dendrogram caluclated for 9 isolates  of \textit{Klebsiella pneumoniae}.
    \item Figure~\ref{fig:segmentation}: Segmentation  for  sample  patches.
    \item Figure~\ref{fig:border}: Border correction.
\end{itemize}

